%% file: tmlr.tex
\documentclass[10pt]{article} % For LaTeX2e
    \usepackage[preprint]{tmlr}
    
    \input{math_commands.tex}

    \usepackage{hyperref}
    \usepackage{url}
    \usepackage{graphicx}
    \usepackage{booktabs}

    \title{MechRL: Reinforcement Learning Agents perform Circuit Discovery for Mechanistic Interpretability}
    
    \author{\name Barsat Khadka \email Barsat.Khadka@usm.edu \\
          \addr School of Computing Sciences and Engineering\\
          The University of Southern Mississippi
          }
    
    \def\month{MM}  % Insert correct month for camera-ready version
    \def\year{YYYY} % Insert correct year for camera-ready version
    \def\openreview{\url{https://openreview.net/forum?id=XXXX}} % Insert correct link to OpenReview for camera-ready version

\begin{document}

    \maketitle
    
    \begin{abstract}
    Mechanistic interpretability seeks to explain a model's behaviour by finding its circuit: the sparse subgraph of the model's computation that is causally responsible for it. Automated methods have made this search systematic, but each one starts afresh for every behaviour, and the effort spent finding one circuit does nothing for the next. Circuit discovery has thus been automated, but not amortised. We ask whether circuit discovery can itself be learned. We frame it as a sequential decision problem over the computation graph of GPT-2 small, in which a policy removes edges until it reaches a compact subgraph that preserves the behaviour, guided by a faithfulness reward defined through causal intervention. A single policy trained across twelve behaviours recovers a faithful circuit for each, and once frozen it transfers to behaviours it never saw during training, recovering their known circuits without further search. A short warm-start improves these transferred circuits, returning far smaller ones than training from scratch. While the learned policy does not match a per-behaviour search on circuit size or cost, it shows that circuit discovery is a learnable, transferable procedure rather than a search repeated for every behaviour.
    \end{abstract}
    
    \section{Introduction}
    Mechanistic interpretability seeks to reverse-engineer neural networks and explain how their internal computations give rise to model behaviour \citep{Olah2020}, rather than treating the network as a black box \citep{bereska2024mechanistic, rai2025practicalreviewmechanisticinterpretability, sharkey2025open}. In transformer language models, this effort centres on finding circuits. A circuit is a sparse subgraph of the model's computation, composed of attention heads, MLPs, and the connections between them, that is causally responsible for a particular behaviour \citep{elhage2021mathematical}. Through careful manual analysis, researchers have recovered circuits for a small number of behaviours such as induction \citep{olsson2022context}, indirect-object identification (IOI) \citep{wang2023interpretability}, the greater-than operation \citep{hanna2023how}, and docstring completion \citep{heimersheim2023docstring}, establishing that specific, sparse sets of components implement specific behaviours.
      
    Because manual analysis does not scale, a growing line of work has automated
    it. ACDC \citep{conmy2023towards} greedily prunes the computation graph to a
    minimal subgraph that preserves a target behaviour, while edge-attribution
    patching \citep{syed2024attribution} and its integrated-gradients variant
    \citep{hanna2024have} approximate the effect of ablating each edge in a
     small number of backward passes and retain only the highest-scoring edges. These methods have made circuit discovery systematic and reproducible, yet they
    share a common limitation: each runs an independent, from-scratch procedure
    for every new behaviour, so the effort invested in discovering one circuit
    confers no advantage on discovering the next. A separate line of methods does learn a per-task mask by gradient descent, whether over model weights \citep{cao-etal-2021-low} or over edges of the computation graph \citep{bhaskar2024finding}, yet each mask is fit from scratch for a single behaviour and does not carry over to the next. Circuit discovery has thus been
    automated, but not amortised. To our knowledge, no existing method learns a
    single discovery procedure that transfers to behaviours unseen during training
    
\begin{figure}[t]
    \centering
    \includegraphics[width=\textwidth]{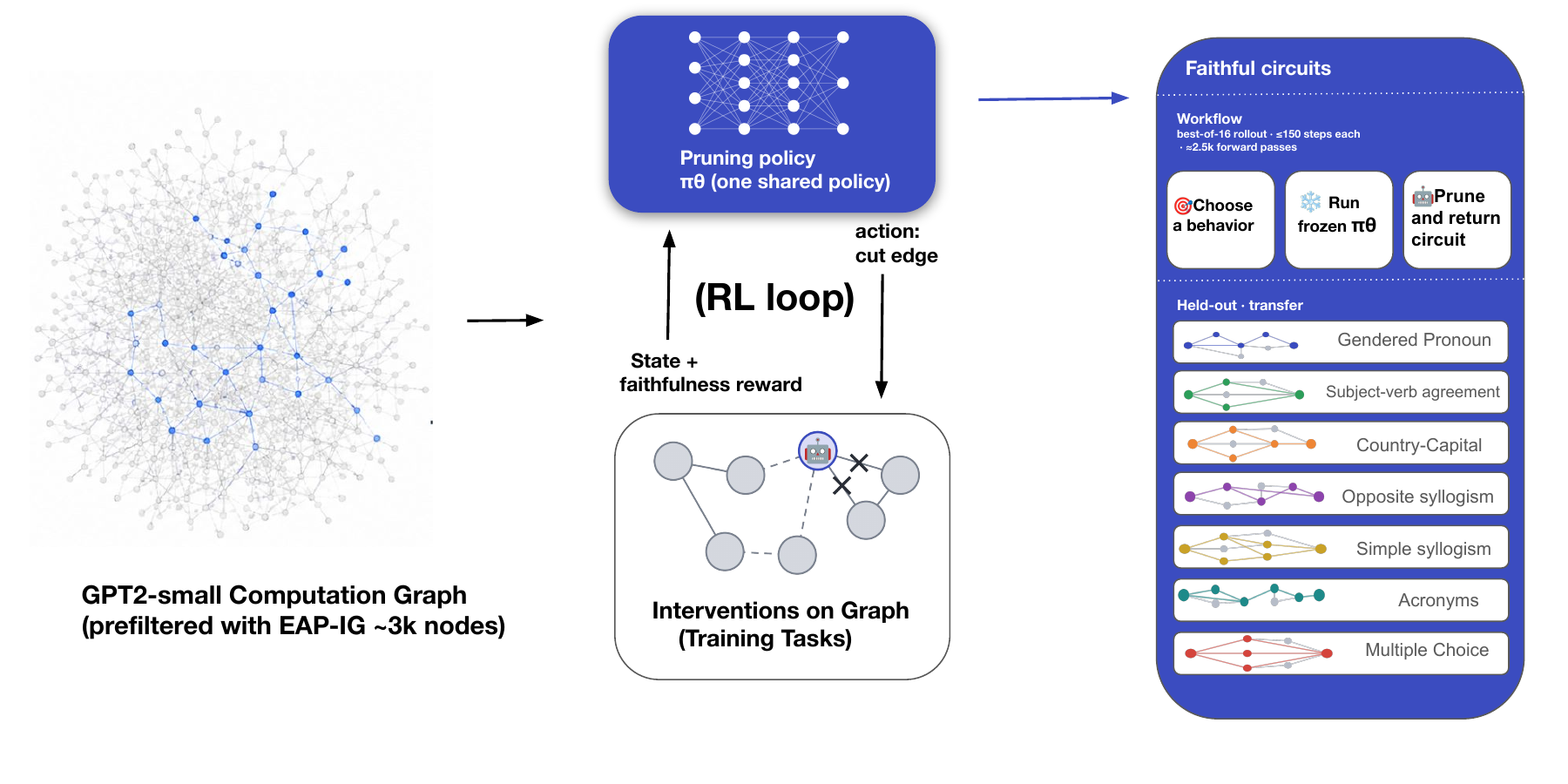}
    \caption{\textbf{Overview of MechRL.} We frame circuit discovery as a sequential decision problem. A single pruning policy $\pi_\theta$ repeatedly removes edges from a prefiltered GPT-2 small computation graph, guided by a faithfulness reward computed through causal patching. Trained jointly across twelve behaviours spanning three families, the same frozen policy recovers a faithful circuit for each, and transfers to a held-out behaviour never seen during training.}
    \label{fig:hero}
    \end{figure}

    In this work, we ask whether circuit discovery can itself be learned. We frame
    edge-level circuit discovery as a sequential decision problem, in which an
    agent observes the computation graph and removes edges until it reaches a
    compact subgraph that preserves the target behaviour, guided by a faithfulness
    reward defined through causal intervention. By training a single policy across
    many behaviours rather than searching afresh for each one, the agent learns a
    reusable discovery procedure, and we show that this procedure transfers to
    behaviours the policy never encountered during training. Figure~\ref{fig:hero} gives an overview of the approach. Our contributions are as follows:
    \begin{itemize}
        \item We formulate edge-level circuit discovery as a reinforcement-learning problem over the model's computation graph, with a faithfulness reward defined through causal intervention.
        \item We show that circuit discovery is a learnable, reusable procedure: a single frozen policy recovers faithful circuits across twelve training behaviours and transfers to held-out behaviours with no per-behaviour search, and a short warm-start returns far smaller circuits than training from scratch.
        \item We show the recovered circuits are genuine rather than reward artefacts or echoes of the prefilter: they recover the canonical components of every behaviour and pass necessity and specificity controls, and the policy keeps causally load-bearing edges its attribution ranking undervalues.
    \end{itemize}
    
    \section{Background}
    \subsection{The mechanistic interpretability workflow}
    Circuit discovery follows a workflow that \citet{conmy2023towards} make explicit and that earlier studies already followed \citep{wang2023interpretability, hanna2023how, heimersheim2023docstring}. It has three steps.
    \begin{enumerate}
        \item \textbf{Choose a behaviour.} Pick a behaviour the model exhibits, build a dataset that elicits it, and fix a metric for how strongly the model performs it.
       \item \textbf{Define the computation graph.} Fix a granularity such as attention heads and MLP layers. This casts the model as a graph whose nodes are components and edges as connections between them.
           \item \textbf{Prune to the circuit.} Iteratively patch away the components the behaviour does not need, checking after each removal that the metric still holds, until a sparse subgraph (circuit) remains. 
    
    \end{enumerate}

    The first two steps are settled once per behaviour. Step 3 is an iterative search over tens of thousands of edges, and it is the step that automated methods such as ACDC \citep{conmy2023towards} and edge attribution \citep{syed2024attribution, hanna2024have} replace. Our agent targets the same step and produces the same kind of output, a sparse faithful subgraph, so its circuits can be measured directly against theirs. We automate the search alone, not the interpretation of what each component computes.
    
    \subsection{Measuring faithfulness}\label{sec:faith}
    \textbf{Computational graph.} Any transformer language model can be cast as a computation graph whose nodes are its components, the attention heads and MLP layers, joined by edges that carry one component's output to another's input. GPT-2 small, the model we study, has 144 attention heads (twelve in each of its twelve layers) and twelve MLP layers \citep{radford2019language}. Because every component writes to a shared residual stream that all later ones read, each node feeds every node downstream of it , giving roughly 32{,}000 edges \citep{elhage2021mathematical, conmy2023towards}. The granularity of this decomposition is a modelling choice, and we take it at the level of attention heads and MLP layers.
    
    \textbf{Patching.} Components outside the circuit are switched off by activation patching \citep{meng, zhang2024towards}. Each behaviour has a clean input that elicits it and a corrupted input that does not. For IOI the clean input ``When John and Mary went to the store, Mary gave a drink to'' is completed with ``John'', and the corrupted input swaps a name so the answer changes \citep{wang2023interpretability}. We run on the clean input but overwrite every edge outside the circuit with its corrupted-run activation, so the circuit is judged on what it alone can reconstruct.
    
    \textbf{Faithfulness.} A circuit is faithful if patching the rest of the model away barely changes its prediction \citep{conmy2023towards, hanna2024have, wang2023interpretability}. We compare two runs on the clean input, the full model and the patched circuit, and measure the gap between their next-token distributions with the Kullback-Leibler (KL) divergence at the prediction position. A faithful circuit gives a small divergence. We normalise it into a faithfulness score,
\begin{equation}
f = 1 - \frac{\mathrm{KL}}{\mathrm{KL}_{\text{cut}}},
\end{equation}
where $\mathrm{KL}_{\text{cut}}$ is the divergence with every edge cut, so $f = 1$ for a perfect match and $f = 0$ for the empty circuit, with higher $f$ more faithful. This $f$ is what the reward and all later results utilize in our setup.
    
    \section{Methodology}
    We cast circuit discovery as a Markov decision process and learn a single policy that solves it across all of our behaviours. The rest of this section gives the formulation, the faithfulness reward, the policy network, and the multi-task training procedure.
    
    \subsection{Circuit discovery as a Markov decision process}
    Starting from a candidate graph with every edge present, the agent removes edges one decision at a time, observing the remaining graph and its faithfulness after each, until it chooses to stop, and the subgraph left standing is the returned circuit. Formally this is a Markov decision process $(\mathcal{S}, \mathcal{A}, T, r, \gamma)$, and we learn the policy $\pi_\theta(a_t \mid s_t)$ to maximise the expected return
    \begin{equation}
    \max_\theta \; \mathbb{E}_{\pi_\theta}\!\left[\sum_{t \ge 0} \gamma^t \, r_t\right],
    \end{equation}
    optimised with PPO \citep{schulman2017ppo} as described below.
    
    \textbf{Computation graph.} We use the edge-level computation graph of \citep{hanna2024have}, whose nodes are GPT-2 small's attention heads and MLPs together with the token input and the logits ($158$ in all), and whose edges carry a node's output to the input of a later node through the residual stream, with a head's query, key, and value counted as separate inputs. This gives roughly $32{,}000$ edges. For each behaviour we rank the edges by EAP-IG attribution and keep the top $K = 3000$ edges, over $M$ distinct parent components, as the candidate set the agent prunes.
    
    \textbf{State ($\mathcal{S}$).} At each step the agent observes the state $s_t = (X^e, X^n, m_t, g_t)$, consisting of edge and node features of the top-$K$ candidate set, the current pruning mask, and a small set of global scalars respectively. The edge features $X^e \in \mathbb{R}^{K \times 16}$ and node features $X^n \in \mathbb{R}^{M \times 7}$ are computed once per behaviour and stay fixed throughout an episode; Table~\ref{tab:state} lists them in full. The pruning mask $m_t \in \{0,1\}^K$ records which candidate edges are still present at step $t$, flipping an entry from one to zero whenever an edge is cut, and the globals $g_t \in \mathbb{R}^6$ track the agent's progress.
    
    \begin{table}[t]
    \centering
    \small
    \caption{State features; each is defined in full in Appendix~\ref{app:features}. Every feature is normalised within each behaviour.}
    \label{tab:state}
    \begin{tabular}{@{}lp{0.8\textwidth}@{}}
    \toprule
    Group & Features \\
    \midrule
    Edge $X^e\,(K{\times}16)$ & signed attribution score; importance rank; parent and child layer; layer distance; parent and child type (one-hot over attention, MLP, input, logits); query/key/value channel (one-hot) \\
    \addlinespace[2pt]
    Node $X^n\,(M{\times}7)$ & layer; type (one-hot); out-degree (number of candidate edges from the node); aggregate (summed) edge score \\
    \addlinespace[2pt]
    Globals $g_t\,(\mathbb{R}^6)$ & step fraction; current faithfulness $f_t$; fraction of edges still present; faithfulness change over the last step; faithfulness of the full candidate set; headroom $f_t-\tau$ (current minus target faithfulness) \\
    \bottomrule
    \end{tabular}
    \end{table}

    \textbf{Actions ($\mathcal{A}$).} At each step, the agent has three choices. It can \emph{stop} and return the current circuit, make a \emph{batch cut} that removes $N$ edges at once with $N$ from $\{1, 3, 10, 30\}$, or make a \emph{node cut} that removes every remaining edge out of one component. Cuts are permanent and a removed edge is never restored.
    
    \textbf{Transition ($T$).} The transition is how the state changes after an action. When the agent performs an action (cuts edges), the environment removes them (with activation patching) and runs one forward pass to measure the new faithfulness score ($f_{t+1}$), and form the next state. The transition is deterministic, so the same action in the same state always gives the same next state. A batch cut of $N$ edges is measured with one run of the model, not $N$, so the agent evaluates the model far fewer times than methods that tests edges one at a time. An episode ends when the agent chooses stop or reaches a fixed limit on the number of steps.

\subsection{Reward ($r$)}\label{sec:reward}
The agent's goal is to find the minimal circuit that explains the behaviour, as measured by faithfulness. We model this objective as a reward function built on a potential $\Phi$ over circuits,
\begin{equation}
\Phi(f, k) = w\left(1 - \frac{k}{K}\right) - \lambda \, \max(0,\, \tau - f),
\end{equation}
where $f$ is the current faithfulness, $k$ the number of retained edges, $K$ the candidate-set size, $\tau$ the faithfulness target, and $w$ and $\lambda$ the weights of the two terms. The first term rewards sparsity and the second is a hinge penalty that activates only when $f$ falls below $\tau$. We set $\lambda$ large relative to $w$, so $\tau$ acts as a near-hard constraint. The reward at each step is the increment in this potential,
\begin{equation}
r_t = \Phi(f_{t+1}, k_{t+1}) - \Phi(f_t, k_t),
\end{equation}
with a fixed penalty of $-0.01$ for cutting an already-removed edge and zero for stopping. Rewarding the change in $\Phi$ rather than $\Phi$ itself is potential-based reward shaping \citep{ng1999policy}, which provides a dense per-step signal without changing the optimal circuit.

\textbf{Per-behaviour target.} We set the faithfulness target $\tau$ per behaviour. The full model is perfectly faithful, but the agent operates within the prefiltered candidate set, whose maximal faithfulness, the \emph{ceiling} $c_i$, lies below the full model's and varies across behaviours. We therefore define $\tau_i = c_i - \delta$ for a small margin $\delta$. A single shared target would be unattainable for low-ceiling behaviours and trivially satisfied for high-ceiling ones, leaving the objective incomparable across behaviours of differing difficulty.

 \subsection{Policy network}
   The policy and value function share a single network that maps a state to a distribution over actions and a value estimate. The candidate-set size $K$ and the number of components $M$ vary across behaviours, so the network scores each edge and each component independently with shared weights rather than through a fixed-width output layer. Its parameter count is therefore independent of $K$ and $M$, and one network applies to every behaviour.
    
    \textbf{Trunk.} Each candidate edge $i$ is embedded by a two-layer MLP from its feature row $X^e_i$ and alive flag $m_{t,i}$, giving $h^e_i = \mathrm{MLP}_e([X^e_i, m_{t,i}]) \in \mathbb{R}^{H}$, and each component $j$ is embedded likewise from $X^n_j$ and the fraction $\bar m_{t,j}$ of its edges still alive, giving $h^n_j \in \mathbb{R}^{H}$, where $H$ is the hidden width. A context vector then pools these embeddings with the globals,
    \begin{equation}
    c = \mathrm{MLP}_c\!\big(\big[\, \overline{h^e},\; \textstyle\max_i h^e_i,\; \overline{h^n},\; \max_j h^n_j,\; g_t \,\big]\big) \in \mathbb{R}^{H},
    \end{equation}
    where $\overline{h^e}$ and $\max_i h^e_i$ are the mean and elementwise maximum over the edge embeddings, and likewise over the components.
    
 \textbf{Heads.} We utilize four heads to turn the trunk's outputs into an action distribution and a value estimate. The \emph{type} head reads the context $c$ and produces a distribution over which kind of action to take (stopping, a batch cut, or a node cut) The \emph{edge} and \emph{node} heads choose the target of the cut, scoring each candidate edge from $[h^e_i, c]$ and each component from $[h^n_j, c]$, with already-cut edges and exhausted components masked out. The \emph{value} head reads $c$ alone and returns the value estimate $V(s_t)$.

    \textbf{Action probability.} An action's probability is the product of the head probabilities along the choices it makes. Stopping has probability $p_\text{type}(\text{stop})$, and a node cut $p_\text{type}(\text{node})\,p_\text{node}(j)$. A batch cut of size $N$ is sampled one edge at a time, re-scoring the remaining edges after each pick so the agent can avoid removing two edges that only matter together. Its log-probability sums the type choice and the conditional edge picks,
    \begin{equation}
    \log \pi(\text{batch}) = \log p_\text{type}(N) + \sum_{l=1}^{N} \log p_\text{edge}\big(e_l \mid e_{1:l-1}\big).
    \end{equation}
 Re-scoring between picks adds little overhead, since the per-edge and per-component embeddings are computed once and only the pooled context is recomputed after each pick.
    
 \subsection{Training}
    We train the single policy on all twelve behaviours jointly with PPO \citep{schulman2017ppo} and generalised advantage estimation \citep{schulman2018high}. Each episode samples a behaviour, so a rollout mixes behaviours and the shared network is updated on all of them, over a frozen copy of GPT-2 small. We de-conflict per-behaviour gradients with PCGrad \citep{pcgrad}. Hyperparameters are listed in the appendix.

    \section{Experimental Setup}\label{sec:setup}
All experiments use GPT-2 small \citep{radford2019language}. We use twelve training behaviours, four variants each of three families (IOI, the greater-than operation, and docstring completion), and hold out seven further behaviours to test transfer. Within a family, the four variants realise one mechanism under different surface forms, varying the sentence frame or the function signature, with their construction detailed in Appendix~\ref{app:tasks}. This variation probes whether a circuit generalises across realisations of a behaviour rather than overfitting a single phrasing, and it provides a within-family transfer setting alongside the cross-family one. For each behaviour we generate twenty clean and corrupted example pairs. Faithfulness is the KL score $f$ of Section~\ref{sec:faith} for every behaviour; the per-behaviour metric in Table~\ref{tab:tasks} is the behaviour's own success measure, used to rank edges in the prefilter (to obtain the top K candidate set) and in the validity controls. The ceiling $c_i$ in the table is the faithfulness of the full prefiltered candidate set, the most a circuit drawn from it, and hence the agent, can reach across variants.

\begin{table}[t]
\centering
\small
\caption{Behaviours used in our experiments for training and testing.}
\label{tab:tasks}
\begin{tabular}{@{}l p{0.40\textwidth} l l c@{}}
\toprule
Behaviour & Example prompt & Output & Metric & $c_i$ \\
\midrule
\multicolumn{5}{@{}l}{\emph{Training (four variants each)}} \\
IOI & ``When Alice and Bob went to the store, Bob gave a book to'' & Alice & logit diff & 0.93--0.99 \\
\addlinespace[2pt]
Greater-than & ``The war lasted from 1812 to 18'' & \textgreater 12 & prob diff & 0.95--0.98 \\
\addlinespace[2pt]
Docstring & {\footnotesize\ttfamily def f(self, a, b, c, d):\newline~~~~"""summary\newline~~~~:param a:\newline~~~~:param b:\newline~~~~:param} & c & logit diff & 0.85--0.92 \\
\midrule
\multicolumn{5}{@{}l}{\emph{Held-out (transfer)}} \\
Gendered pronoun & ``So Dave is a really great friend, isn't'' & he & logit diff & 1.00 \\
\addlinespace[2pt]
Subject-verb & ``The author near the cars'' & is & logit diff & 0.97 \\
\addlinespace[2pt]
Acronyms & ``The Federal Reserve Board (FR'' & B & logit diff & 0.99 \\
\addlinespace[2pt]
Simple syllogism & ``Statement A is true. Statement B matches statement A. Statement B is'' & true & logit diff & 0.93 \\
\addlinespace[2pt]
Opposite syllogism & ``Statement A and B are opposite. A is true. B is'' & false & logit diff & 0.92 \\
\addlinespace[2pt]
Country-capital & ``The capital of France is'' & Paris & logit diff & 0.92 \\
\addlinespace[2pt]
Multiple choice & ``Answer Choices: (A) cat (B) dog Answer: ('' & A & KL & 0.95 \\
\bottomrule
\end{tabular}
\end{table}

\subsection{Metrics and validity}
We report three quantities for each circuit, its faithfulness $f$, its minimality $|C|$ (the number of edges it retains) and its cost (the number of forward passes used to find it). Faithfulness and minimality trade off against each other, so we compare circuits at matched faithfulness, where a smaller and cheaper circuit is better.

\textbf{Validity battery.} A circuit can reach high faithfulness yet still be an artefact of the reward, as a form of reward hacking \citep{skalse2022defining}. To test that a discovered circuit is genuine we check three conditions. \emph{Ground-truth recovery} asks whether the circuit contains the components prior human analysis identified for the behaviour; we measure the fraction recovered against the published circuits where they exist, and since those were found by people rather than by our reward, recovering them shows the circuit captures the real mechanism. \emph{Necessity} asks whether the behaviour genuinely depends on the circuit; ablating the circuit from the full model collapses the behaviour, confirming the circuit is load-bearing rather than incidental. \emph{Specificity} asks whether the identity of the edges matters, not merely their count; substituting a random subgraph of the same size yields an unfaithful circuit, so the result is attributable to these particular edges rather than to retaining enough of them.

    \section{Results}

\subsection{Transfer to held-out behaviours}\label{sec:transfer}

\begin{figure}[t]
\centering
\includegraphics[width=\textwidth]{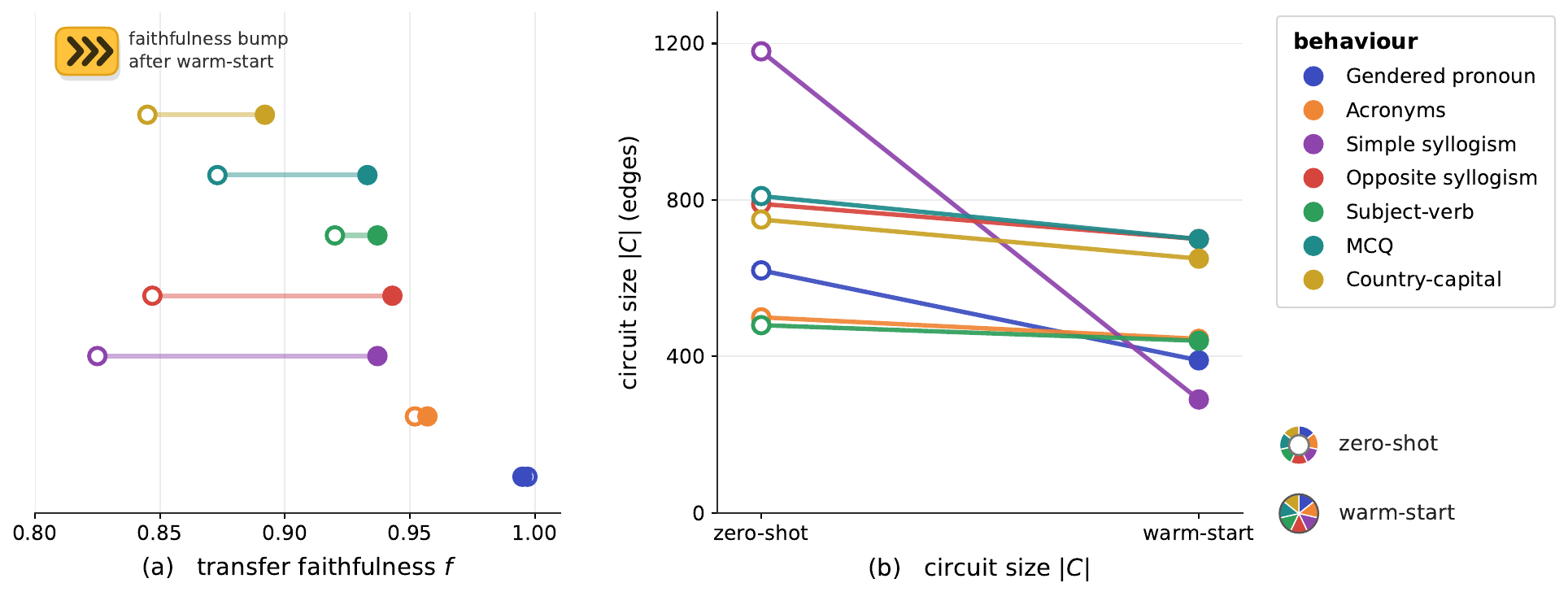}
\caption{Transfer of the frozen twelve-task policy to seven held-out behaviours, zero-shot (open markers) versus a short warm-start (filled). (a) Faithfulness $f$ and (b) circuit size $|C|$ in edges. Warm-start raises faithfulness and shrinks every circuit.}
\label{fig:transfer}
\end{figure}

\textbf{Zero-shot transfer.} We test whether the trained policy has acquired a transferable procedure for circuit discovery rather than a set of behaviour-specific solutions. We freeze the policy trained on the twelve behaviours of Section~\ref{sec:setup} and apply it, without further training, to seven held-out behaviours: six whose circuits in GPT-2 small have been established by prior work \citep{Acronyms, Syllogism, mcq, subject-verb, genderedpronoun, causal}, and country--capital factual recall, a different mechanism family with no published sparse circuit \citep{meng}, as a harder transfer test. The policy is run as during training, drawing a small number of rollouts under its frozen weights and retaining the most faithful, with no gradient updates and no behaviour-specific tuning. Each returned circuit is scored with the faithfulness $f$ of Section~\ref{sec:faith} and the validity battery of Section~\ref{sec:setup}, and for behaviours whose responsible heads are named in prior work we report the fraction of those heads the circuit recovers. Figure~\ref{fig:transfer} summarises the result, plotting each behaviour twice, once for the zero-shot circuit (open markers) and once after the short warm-start described below (filled markers); Table~\ref{tab:transfer} gives the zero-shot numbers in full.

\begin{table}[t]
\centering
\small
\caption{Transfer of the frozen twelve-task policy to seven held-out behaviours, zero-shot and after a short warm-start. The necessity, specificity and component-recovery columns are for the zero-shot circuit. $^\dagger$KL-attribution prefilter.}
\label{tab:transfer}
\begin{tabular}{lccccccc}
\toprule
 & \multicolumn{2}{c}{zero-shot} & \multicolumn{2}{c}{warm-start} & & & \\
\cmidrule(lr){2-3}\cmidrule(lr){4-5}
Behaviour & $f$ & $|C|$ & $f$ & $|C|$ & necessity & specificity & components \\
\midrule
Gendered pronoun~\citep{genderedpronoun}        & 0.997 & 628  & 0.995 & 390 & 0.00    & 0.00 & 7/7          \\
Subject-verb agreement~\citep{subject-verb}  & 0.920 & 480  & 0.937 & 440 & 0.00    & 0.00 & 11/12    \\
Acronyms~\citep{Acronyms}                & 0.956 & 491  & 0.957 & 445 & 0.00    & 0.00 & 7/8          \\
Simple syllogism~\citep{Syllogism}        & 0.826 & 1175 & 0.920 & 290 & 0.04    & 0.00 & 5/5          \\
Opposite syllogism~\citep{Syllogism}      & 0.847 & 779  & 0.943 & 700 & $-0.05$ & 0.00 & 9/9 \\
Country-capital~\citep{meng}         & 0.843 & 753  & 0.892 & 650 & 0.00    & 0.00 & ---          \\
Multiple choice~\citep{mcq}$^\dagger$ & 0.874 & 819 & 0.933 & 700 & $-0.01$ & 0.00 & 3/3 \\
\bottomrule
\end{tabular}
\end{table}

\textbf{Few-shot adaptation (warm start).} We next ask whether brief behaviour-specific training improves on the zero-shot result. Warm-starting from the frozen policy, we resume training on a single held-out behaviour for eighty iterations and re-evaluate the circuit it returns. Across all seven behaviours the warm-started circuit is at least as faithful as its zero-shot counterpart, and strictly more faithful wherever the zero-shot circuit had not already saturated (Figure~\ref{fig:transfer}a). Circuit size falls as well, but for most behaviours the change is modest, so faithfulness improves while the number of edges stays close to where it began, which we read as the policy filtering the same circuit more cleanly rather than finding a smaller one (Figure~\ref{fig:transfer}b). The simple syllogism is the clear exception, where the circuit is cut sharply from $1175$ to $290$ edges while its faithfulness climbs from $0.826$ to $0.92$. Taken together these results show that a short warm-start of eighty iterations reliably improves a transferred circuit, and we therefore treat it as a cheap refinement step on top of zero-shot transfer. The necessity and specificity columns report the faithfulness $f$ of the knocked-out and random same-size circuits of Section~\ref{sec:setup}, so a value near zero is the desired result, and a small negative value only reflects that $f$ is not clamped below zero.

\begin{figure}[t]
\centering
\includegraphics[width=\textwidth]{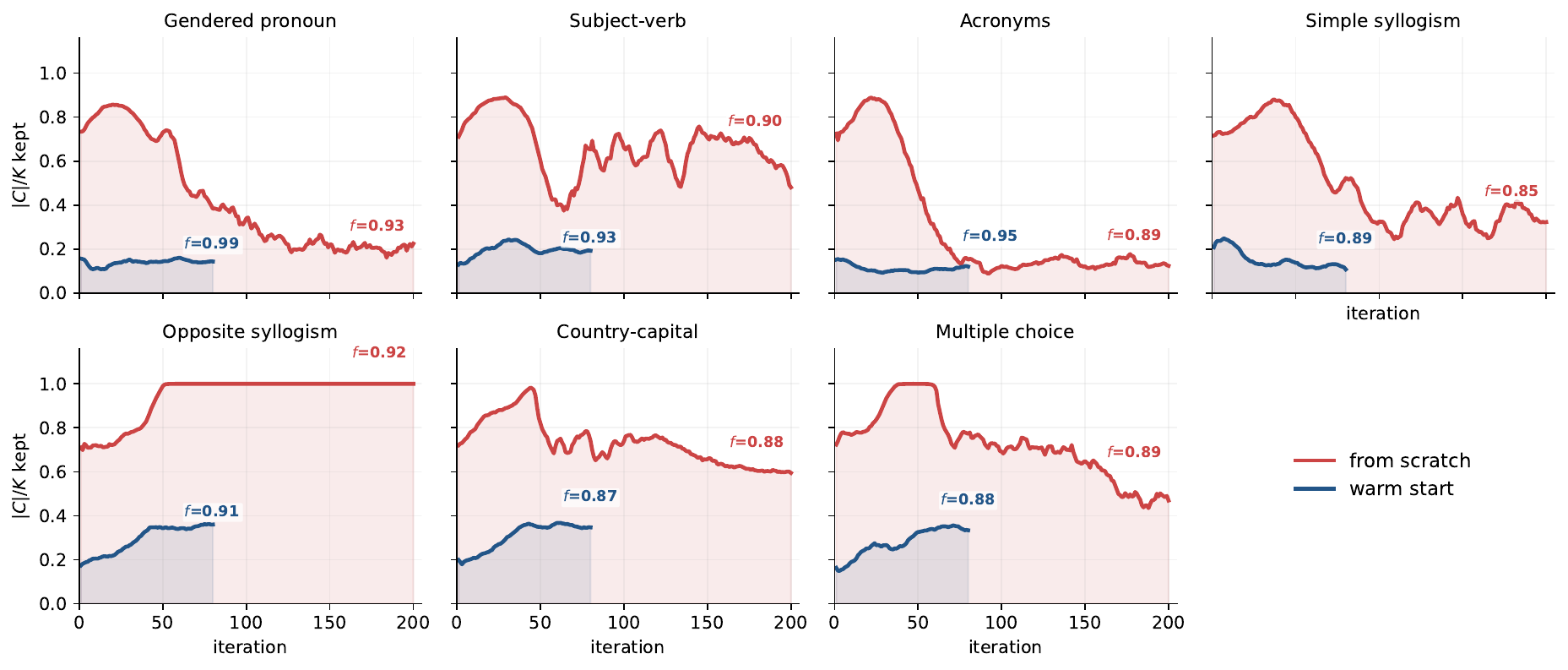}
\caption{Fraction of the candidate set retained, $|C|/K$, per iteration on each held-out behaviour: warm-started from the frozen twelve-task policy (blue) versus trained from random initialisation (red); $f$ is the final faithfulness of each run. At matched faithfulness the from-scratch policy keeps a far larger circuit and prunes it down only slowly, while warm-start is small from the first iterations.}
\label{fig:warmscratch}
\end{figure}

\textbf{Warm start versus training from scratch.} The warm-start gain is only evidence of transfer if it comes from the pretrained policy and not from the held-out behaviour being easy to fit from any initialisation. We therefore compare each warm-started run against a policy trained from random initialisation on the same behaviour, holding the reward, the candidate set, the faithfulness target and every hyperparameter fixed so that the initialisation is the only difference, and we train the from-scratch policy for more than twice as long, two hundred iterations against eighty. Faithfulness cannot separate the two, because the prefilter already makes the candidate set highly faithful: any sufficiently large subgraph clears the target, so both policies reach it within a few tens of iterations and reaching it says little. They differ instead in how much they prune, and minimality is where they separate. From its first iterations the warm-started policy returns circuits of a few hundred edges, while the from-scratch policy begins by keeping most of the candidate set and prunes it down only slowly; after two hundred iterations its circuits remain between $1.3$ and $3.3$ times larger, a median of $1.65$, at comparable or lower faithfulness on six of the seven behaviours, and on the seventh it is no smaller while less faithful (Figure~\ref{fig:warmscratch}). The warm-started policy is small from the start because it already holds a pruning procedure, whereas the from-scratch policy must discover one. This is the most direct evidence that what transfers is a general pruning skill rather than a set of behaviour-specific circuits.

\subsection{Does the policy learn beyond the prefilter?}\label{sec:learn}

\begin{figure}[t]
\centering
\includegraphics[width=\textwidth]{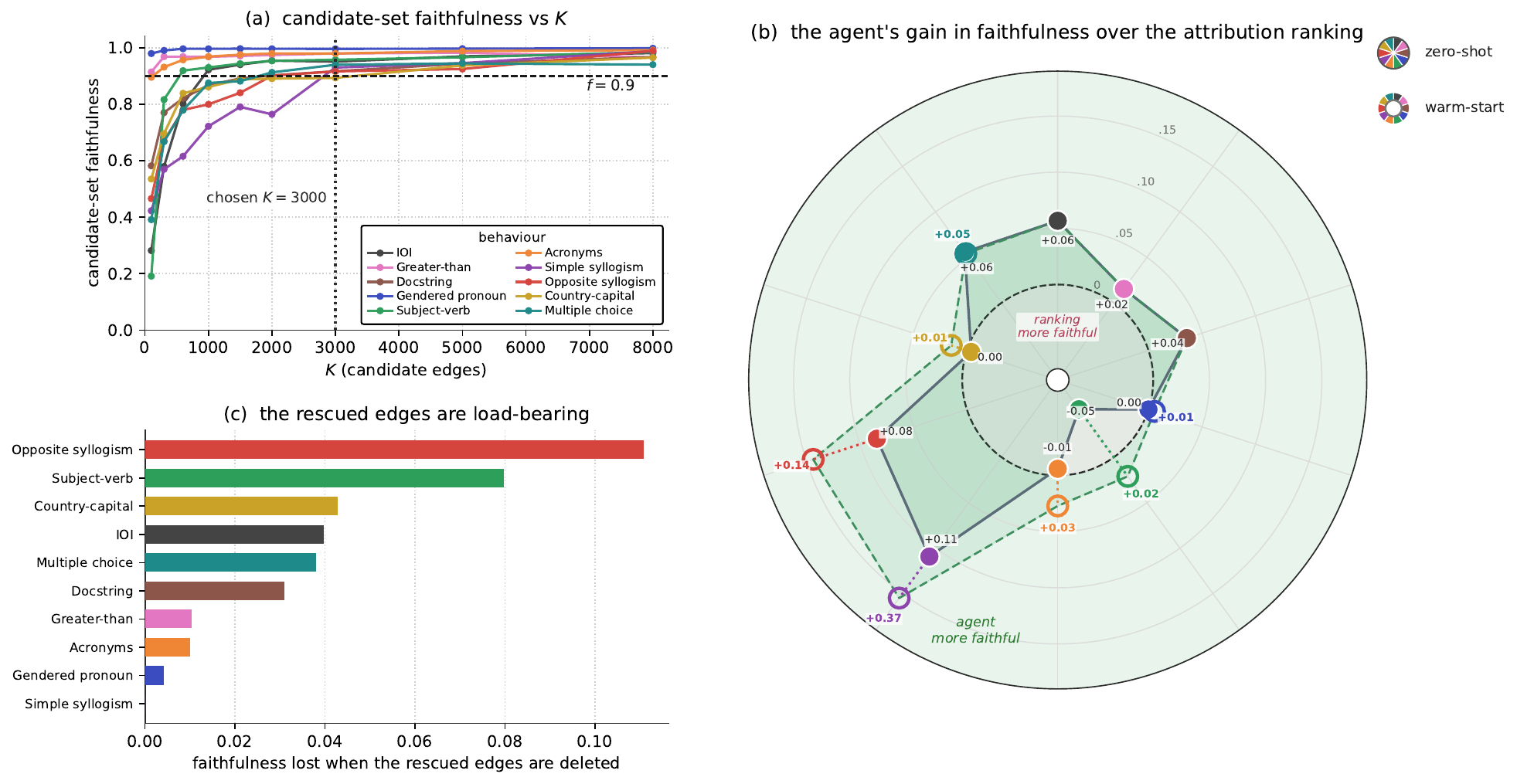}
\caption{Does the policy learn beyond the prefilter? Colour denotes the behaviour (legend in (a)). \textbf{(a)} Candidate-set faithfulness versus the prefilter size $K$. \textbf{(b)} The agent's gain in faithfulness over the size-matched top-$|C|$-by-score circuit (radius, with the value at each dot); the green ring marks where the agent is more faithful than the ranking, and for the held-out behaviours an open marker joined by a dotted spoke gives the gain after the short warm-start. \textbf{(c)} Faithfulness lost when each behaviour's below-cutoff (\emph{discordant}) edges are deleted.}
\label{fig:learn}
\end{figure}

The agent never searches the full computation graph. A prefilter first scores every edge by EAP-IG attribution \citep{syed2024attribution, hanna2024have} and keeps the top $K$ as the candidate set the agent acts on. We take $K = 3000$: it is the smallest candidate set whose faithfulness ceiling reaches $0.9$ for every behaviour, including the hardest (Figure~\ref{fig:learn}a), while keeping the action space tractable. The scoring is EAP-IG's, and keeping the top $K$ by that score is its simplest selection rule \citep{hanna2024have}, which we apply at a deliberately large $K$ to form a candidate pool rather than a final circuit; EAP-IG instead extracts its circuit directly from the same scores by a \emph{greedy search}, where we hand the pool to the RL agent. The sceptical reading is therefore that the agent inherits an already-faithful set and adds little of its own. We test this by turning the tools of mechanistic interpretability on the discovery policy itself rather than on the model it studies, asking whether the policy improves on the attribution that built its candidate set, whether it keeps mechanism that attribution alone would discard, and whether its circuits agree with the components found by hand. Its minimisation is genuine, even if it does not match the efficiency of EAP-IG's greedy search (Section~\ref{sec:comparison}).

\textbf{The policy improves on its own candidate set.} The agent prunes the top-$K$ candidate set down to $|C|$ edges. The question is whether those are simply the $|C|$ highest-scoring edges, the ranking it was handed, or a selection of its own. They are its own, and more faithful (Figure~\ref{fig:learn}b): at matched size the agent's circuit is at least as faithful as the top-$|C|$ by score on nine of the ten behaviours, with the gain largest on the diffuse behaviours where attribution misranks most and the two level only where the score-ranked circuit already sits near the ceiling. Subject--verb agreement is the lone zero-shot exception, and the short warm-start of Section~\ref{sec:transfer} lifts even it above the ranking while widening the advantage on every other behaviour. The agent is therefore selecting edges for the behaviour, not echoing the score, most where the attribution is least reliable.

\textbf{The policy keeps edges that attribution undervalues.} EAP-IG's own analysis shows its score correlates only weakly with an edge's true causal effect \citep{hanna2024have}, so the top-$|C|$ by score leaves out some edges the behaviour actually needs. The agent reaches past the cutoff to keep them. We call its kept edges that rank below the top-$|C|$ \emph{discordant}, the ones a score-ranked circuit of the same size would have dropped. They are thirty to fifty per cent of its circuit on every behaviour, well below the sixty-three to eighty-five per cent a random subset would hold, so the agent follows the ranking for most of its edges while still rescuing a consistent minority from beneath it. Those rescued edges carry the behaviour: deleting them lowers faithfulness on nine of the ten behaviours (Figure~\ref{fig:learn}c), the simple syllogism the lone exception. The agent keeps low-scored edges the attribution undervalues and the behaviour depends on.

\textbf{The circuits agree with hand-found mechanism.} A faithful circuit can still be an artefact of the reward rather than the real mechanism \citep{skalse2022defining}. As a check external to our objective we measure how many of the canonical components prior work identified for each behaviour, its attention heads together with any MLPs the prior work names, the circuit contains (the components column of Table~\ref{tab:transfer}): across the six held-out behaviours with named components, the circuits recover most or all, from $7/7$ for the gendered pronoun to $11/12$ for subject--verb agreement, while the necessity and specificity columns of the same table show that the behaviour collapses when the circuit is ablated and that a random subgraph of equal size is unfaithful. Recovering components found by people, not by our reward, indicates the circuits capture the behaviour's actual mechanism rather than a high-faithfulness artefact.

\subsection{Comparison with per-behaviour search}\label{sec:comparison}

\begin{figure}[t]
\centering
\includegraphics[width=\textwidth]{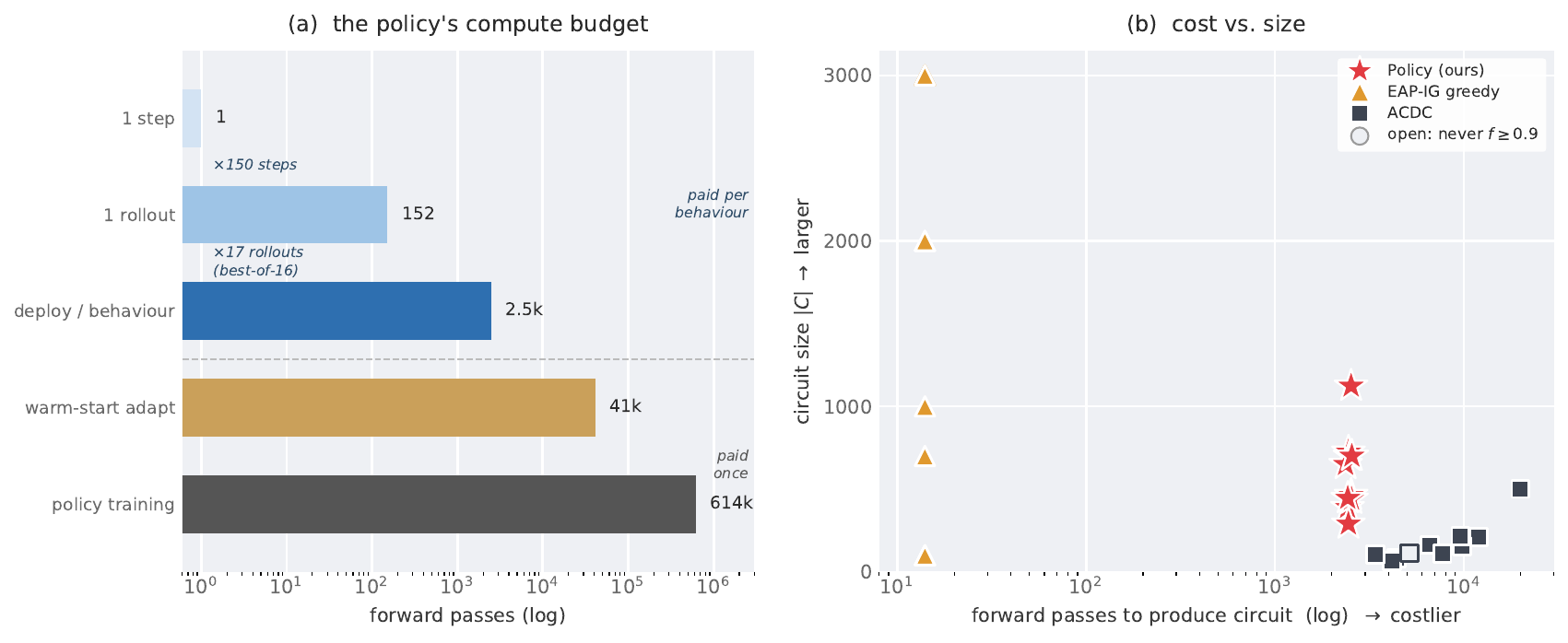}
\caption{\textbf{(a)} The policy's own compute budget: one best-of-sixteen deployment is $17$ rollouts $\times\,{\sim}150$ steps $\approx 2.5$k forward passes, paid per behaviour, above a one-time warm-start adaptation ($\sim$40k per held-out behaviour) and the one-time policy training ($\sim$$6\times10^5$). \textbf{(b)} The cost--minimality plane: forward passes to produce a circuit against its size at $f\ge0.9$, one marker per behaviour. ACDC (small, expensive) and EAP-IG greedy (cheap, but size-scattered) define the Pareto frontier; the policy sits in the dominated interior. Open markers never reach $f=0.9$.}
\label{fig:cost}
\end{figure}

We place the policy against the two standard families of per-behaviour circuit discovery on the same computation graph and the same faithfulness measure $f$: ACDC \citep{conmy2023automated} and EAP-IG greedy \citep{hanna2024have}. Both are re-run from scratch for every behaviour; the policy is run once per behaviour as a fixed best-of-sixteen rollout under frozen (training) or warm-started (held-out) weights. The three methods occupy three points on the cost--minimality trade-off (Table~\ref{tab:acdc}, Figure~\ref{fig:cost}).

\textbf{The policy is costly and it underperforms, but discovery is learnable.} On the two axes a circuit-discovery method is judged on, the forward passes it spends and the size of the circuit it returns at a fixed faithfulness, the policy is not the best. ACDC returns smaller circuits on nine of ten behaviours and EAP-IG greedy is far cheaper, so in the cost--minimality plane the policy sits in the interior rather than on the frontier those two define (Figure~\ref{fig:cost}b). The policy is a middle point, cheaper than ACDC and smaller than attribution-greedy on most behaviours, but middle is neither minimal nor cheapest; its deployment costs about twenty-five hundred passes (Figure~\ref{fig:cost}a), and the warm-start that yields its smallest circuits adds roughly forty thousand more per held-out behaviour, above what ACDC spends. What the policy offers that neither baseline can is that the discovery procedure is learned once and then reused: seven of the ten behaviours in Table~\ref{tab:acdc} are held-out, and their circuits come from the single frozen policy of Section~\ref{sec:transfer} with no per-behaviour search, where ACDC and EAP-IG re-run their full search every time. We claim no saving in total compute, since training the policy costs of order $6\times10^5$ passes; the contribution is that circuit discovery is a learnable process, one procedure that amortises across behaviours and transfers to ones it never saw.

\textbf{ACDC is the minimal baseline, at a price.} ACDC prunes by a threshold $\tau$: it removes every edge whose ablation changes the output by less than $\tau$, so a smaller $\tau$ keeps more edges and a larger one keeps fewer. We sweep $\tau \in \{0.001, 0.003, 0.01\}$ and take, for each behaviour, the smallest circuit that still reaches the same target faithfulness $f\!\ge\!0.9$ the policy is held to. Minimising circuit size at a faithfulness target is exactly what this per-edge search optimises, and on nine of ten behaviours it returns the smallest faithful circuit, often several times smaller than the policy's. It pays for this in cost: judging each edge by ablating it and running the model costs several thousand to nearly twenty thousand forward passes per behaviour, and the search begins again from nothing for every behaviour. It is also not infallible --- on country--capital no threshold reaches $f=0.9$ at all (its best is $0.72$), so even a minimal per-edge search has behaviours it cannot faithfully capture within this graph. That ACDC, at this cost, finds smaller circuits than we do is the honest reading of our result: the policy's circuits are not the minimal explanation of each behaviour, because the policy is not solving for one behaviour's minimal circuit but has learned a single general pruning procedure it applies across all of them.

\textbf{Attribution-greedy is cheap but proxy-bound.} EAP-IG greedy selects edges from precomputed scores, so its selection is effectively free, and on sharply localised behaviours (greater-than, the gendered pronoun, acronyms) it finds faithful circuits of around a hundred edges, \emph{smaller} than the policy's. But where the mechanism is diffuse it degrades sharply: on the syllogisms, country--capital, the docstring and IOI it needs most of the candidate set, and on several it never reaches $f=0.9$ within $K=3000$ at all. Crucially, this is a property of the attribution proxy, not of one scoring choice: switching the target from the logit difference to the KL the circuits are actually scored on (the \emph{ld} and \emph{kl} columns) merely moves \emph{which} behaviours fail rather than fixing them. This is the gap EAP-IG's own analysis documents, that attribution scores correlate only weakly with true causal effect \citep{hanna2024have}.

\begin{table}[t]
\centering
\footnotesize
\caption{The three methods at matched faithfulness $f\!\ge\!0.9$ on the same graph. ACDC: smallest circuit over a $\tau$ sweep ($^{*}$\,never reaches $0.9$; best shown). EAP-IG greedy: size\,$(f)$ under logit-difference (\emph{ld}) and KL (\emph{kl}) attribution ($^{\dagger}$\,never reaches $0.9$ within $K\!=\!3000$); selection is forward-pass-free, hence no passes. Policy: deployed circuit (trained for the three training behaviours, warm-started for the seven held-out) at best-of-sixteen.}
\label{tab:acdc}
\begin{tabular}{l ccr cc ccr}
\toprule
 & \multicolumn{3}{c}{ACDC} & \multicolumn{2}{c}{EAP-IG greedy $|C|\,(f)$} & \multicolumn{3}{c}{Policy (ours)} \\
\cmidrule(lr){2-4}\cmidrule(lr){5-6}\cmidrule(lr){7-9}
Behaviour & $|C|$ & $f$ & passes & \emph{ld} & \emph{kl} & $|C|$ & $f$ & passes \\
\midrule
IOI                & 211          & 0.90 & 11918 & 1000\,(.91)          & 1500\,(.91)          & 1123 & 0.96 & 2528 \\
Greater-than       & 84           & 0.94 & 4383  & 100\,(.93)           & 100\,(.92)           & 464  & 0.98 & 2529 \\
Docstring          & 502          & 0.94 & 19940 & 2000\,(.90)          & 3000\,(.87)$^\dagger$ & 725  & 0.89 & 2448 \\
Gendered pronoun   & 67           & 0.94 & 4155  & 100\,(.98)           & 100\,(.98)           & 390  & 0.99 & 2422 \\
Subject-verb       & 163          & 0.91 & 6590  & 700\,(.92)           & 700\,(.92)           & 440  & 0.94 & 2539 \\
Acronyms           & 103          & 0.94 & 3401  & 100\,(.91)           & 100\,(.90)           & 445  & 0.96 & 2427 \\
Simple syllogism   & 108          & 0.93 & 7682  & 3000\,(.90)$^\dagger$ & 3000\,(.87)$^\dagger$ & 290  & 0.92 & 2446 \\
Opposite syllogism & 155          & 0.92 & 9782  & 2000\,(.90)          & 3000\,(.95)          & 700  & 0.94 & 2494 \\
Country-capital    & 113$^{*}$    & 0.72 & 5171  & 3000\,(.89)$^\dagger$ & 3000\,(.92)          & 650  & 0.89 & 2364 \\
Multiple choice    & 217          & 0.94 & 9599  & 2000\,(.91)          & 2000\,(.91)          & 700  & 0.93 & 2551 \\
\bottomrule
\end{tabular}
\end{table}

    \section{Limitations and Future work}

The main limitation is the prefilter. The full edge graph of GPT-2 small has on the order of $32{,}000$ edges, a discrete action space too large for the agent to search directly, so we reduce it to the top $K=3000$ candidate edges with EAP-IG before the agent acts. This has two consequences. The prefilter already performs part of the discovery, since it discards most of the graph by gradient attribution, so the agent refines a candidate set rather than searching the model from scratch. It also bounds what the agent can reach, because the candidate-set ceiling $c_i$ caps faithfulness and a behaviour whose faithful circuit is spread across many edges needs a large $K$ that the agent cannot act over. Induction is the clearest case. It has a well-studied circuit \citep{olsson2022context}, but in GPT-2 small its faithful circuit is diffuse, and a candidate set of order $16{,}000$ edges was needed before a faithful circuit lay within reach, far beyond a tractable action space. We excluded induction for this reason, and the choice of $K=3000$ throughout is a compromise between coverage and a search space the agent can handle.

Two further limitations follow from the same design. The circuits the agent returns are larger than those a per-behaviour search such as ACDC produces at matched faithfulness, since the agent is trained for amortised transfer rather than minimality and sits at a different point on the faithfulness-size trade-off. And we validate on GPT-2 small, the model for which component-level ground-truth circuits exist, so the same action-space problem makes scaling to larger models a separate challenge rather than a free extension.

\paragraph{Future work.} A natural way to remove the prefilter is to learn a model of the environment the agent searches. The expensive component of every step is the faithfulness evaluation, one forward pass of GPT-2 for each candidate circuit. If a learned model predicted the effect of an edit on faithfulness directly, the agent could plan over the full graph by rolling out simulated edits, in the manner of model-based reinforcement learning and self-play, rather than calling the frozen model at every step. This would let the agent search the complete $32{,}000$-edge graph without a prefilter once an accurate model of the circuit's behaviour has been learned, and is the direction we think most likely to remove the present dependence on attribution-based candidate selection.

    \section{Conclusion}

We framed circuit discovery as a reinforcement learning problem and trained a single policy, jointly over twelve behaviours, to prune the computation graph of GPT-2 small into faithful circuits. The policy transfers to behaviours it never saw, recovering faithful circuits for seven held-out behaviours and the specific heads earlier work identified for them, and a short warm-start improves on this further. Unlike a per-behaviour search, the learned policy is amortised and transferable, which is the property we set out to demonstrate. The main obstacle to scaling the approach is the size of the raw action space, which we presently manage with a prefilter and hope to remove by learning a model of the environment the agent searches.
    \section*{Acknowledgement}

The author sincerely acknowledges the Mississippi Center for Supercomputing Research (MCSR) for providing computational resources and support.

    \bibliography{tmlr}
    \bibliographystyle{tmlr}
    
    \appendix
    \section{Appendix}
\subsection{Behaviour and variant construction}\label{app:tasks}
Each behaviour is generated from a template with slots for its varying tokens, paired with a corrupted template that breaks the behaviour. For IOI the clean template places two names in an ABBA pattern and asks for the indirect object, and the four variants change the surrounding frame, the opener, the connective, or the scenario; the corrupted input shuffles the names so the indirect object changes. For greater-than the clean template states a year span and asks for the second year, and the variants change the frame around the span (\emph{lasted from}, \emph{from \ldots to}, \emph{began \ldots ended}, \emph{took place between}); the corrupted input sets the first year low so any completion is valid. For docstring the clean template lists a function's arguments and documents them in order, and the variants change the signature and cue style, a class method, a free function, bare \texttt{:param} entries, or entries with descriptions; the corrupted input shuffles the argument order so position no longer predicts the next one. Each behaviour uses twenty such pairs.

\subsection{Hyperparameters}
We prefilter each behaviour's computation graph to the top $K = 3000$ edges using EAP-IG attribution with five integration steps. The policy network uses a hidden width of $H = 128$, batch-cut sizes $\{1, 3, 10, 30\}$, and a step budget of $150$ actions per episode. The reward uses minimality weight $w = 1$, threshold penalty $\lambda = 5$, and per-behaviour target $\tau_i = c_i$ (margin $\delta = 0$). We train for $1200$ PPO iterations of $512$ transitions each, with discount $\gamma = 0.99$, GAE parameter $\lambda_{\text{GAE}} = 0.95$, clip coefficient $0.2$, entropy coefficient $0.01$, value coefficient $0.5$, gradient-norm clip $0.5$, four update epochs over four minibatches, and Adam at learning rate $2.5\times10^{-4}$ annealed linearly to zero. Multi-task updates combine per-behaviour gradients with PCGrad.

\subsection{State features in detail}\label{app:features}
The state $s_t = (X^e, X^n, m_t, g_t)$ pairs a static description of the candidate graph ($X^e$, $X^n$, computed once per behaviour) with the dynamic progress of the current episode ($m_t$, $g_t$). Every numeric feature is normalised within each behaviour so that scales are comparable across behaviours of different sizes. We describe each group of Table~\ref{tab:state} in turn; the count after each feature is the number of dimensions it contributes.

\paragraph{Edge features ($X^e \in \mathbb{R}^{K\times16}$), one row per candidate edge.}
\begin{itemize}
  \item \textbf{Signed attribution score} (1): the edge's EAP-IG score, kept signed so the policy sees both the magnitude and the direction of the edge's estimated effect on the metric.
  \item \textbf{Importance rank} (1): the edge's position in the prefilter ordering by $|\text{score}|$, normalised to $[0,1]$, so the policy knows where the edge sits relative to the rest of the candidate set.
  \item \textbf{Parent and child layer} (2): the layer indices of the edge's source and destination components.
  \item \textbf{Layer distance} (1): child layer minus parent layer, i.e.\ how far forward the edge reaches.
  \item \textbf{Parent and child type} (8): two one-hot vectors over \{attention head, MLP, input embedding, logits\} identifying what the source and destination are.
  \item \textbf{Query/key/value channel} (3): a one-hot vector marking which of an attention head's query, key, or value inputs the edge feeds; zero when the destination is not an attention head.
\end{itemize}

\paragraph{Node features ($X^n \in \mathbb{R}^{M\times7}$), one row per component.}
\begin{itemize}
  \item \textbf{Layer} (1): the component's layer index.
  \item \textbf{Type} (4): a one-hot vector over \{attention head, MLP, input embedding, logits\}.
  \item \textbf{Out-degree} (1): the number of candidate edges leaving the component, a proxy for how central it is in the candidate set.
  \item \textbf{Aggregate edge score} (1): the summed $|\text{attribution}|$ of the component's candidate edges, summarising its total estimated importance.
\end{itemize}

\paragraph{Global features ($g_t \in \mathbb{R}^6$), six scalars tracking episode progress.}
\begin{itemize}
  \item \textbf{Step fraction}: the fraction of the step budget used so far, telling the policy how much time remains.
  \item \textbf{Current faithfulness $f_t$}: the faithfulness of the circuit at step $t$.
  \item \textbf{Fraction of edges present}: the number of kept edges divided by $K$.
  \item \textbf{Faithfulness change}: $f_t - f_{t-1}$, the immediate effect of the previous action.
  \item \textbf{Candidate-set faithfulness}: the ceiling $c_i$ reached with all $K$ edges present, a fixed reference for the achievable maximum.
  \item \textbf{Headroom $f_t - \tau$}: current faithfulness minus the per-behaviour target, signalling whether the circuit is above or below the constraint.
\end{itemize}

    \end{document}

%% file: math_commands.tex
\usepackage{amsmath,amsfonts,bm}

\def\eqref#1{equation~\ref{#1}}
\def\1{\bm{1}}

\DeclareMathAlphabet{\mathsfit}{\encodingdefault}{\sfdefault}{m}{sl}
\SetMathAlphabet{\mathsfit}{bold}{\encodingdefault}{\sfdefault}{bx}{n}